\pdfoutput=1

\documentclass[11pt]{article}

\usepackage[]{acl}

\usepackage{times}
\usepackage{latexsym}

\usepackage[T1]{fontenc}

\usepackage[utf8]{inputenc}

\usepackage{microtype}
\usepackage{graphicx}

\usepackage{CJK}
\usepackage{indentfirst}
\usepackage{amsmath}
\usepackage{cases}
\usepackage{subfigure}
\usepackage{multirow}

\title{SimOAP: Improve Coherence and Consistency in Persona-based Dialogue Generation via Over-sampling and Post-evaluation}

\author{
Junkai Zhou$^{1,2}$,
Liang Pang$^{1}$\thanks{\ \ Corresponding authors},
Huawei Shen$^{1,2}$,
Xueqi Cheng$^{1,2}$\footnotemark[1]\\
$^{1}$Data Intelligence System Research Center,\\
 Institute of Computing Technology, CAS\\
 $^{2}$University of Chinese Academy of Sciences \\
{\tt\ \{zhoujunkai20z, pangliang,shenhuawei,cxq\}@ict.ac.cn}}

\begin{document}
\maketitle
\begin{abstract}
Language models trained on large-scale corpora can generate remarkably fluent results in open-domain dialogue. 
However, for the persona-based dialogue generation task, consistency and coherence are also key factors, which are great challenges for language models.
Existing works mainly focus on valuable data filtering, model structure modifying, or objective function designing, while their improvements are limited and hard to generalize to all types of pre-trained language models. 
However, we find that language models can produce consistent and coherent responses if we consider enough generations. 
Thus, the problems lay in large-scale response generation and target response selection.
In this work, a simple but effective two-stage SimOAP strategy is proposed, i.e., over-sampling and post-evaluation. 
The over-sampling stage takes large-scale responses from existing trained models efficiently via off-the-shelf distilling and compressing methods, and the post-evaluation stage selects a good response based on multiple well-designed evaluation metrics from large-scale candidates.
Experimental results show that the proposed plug-in SimOAP strategy improves the backbone models and outperforms the baseline strategies in both automatic and human evaluations.

\end{abstract}

\section{Introduction}

\begin{figure}
    \centering 
    \includegraphics[width=7.7cm]{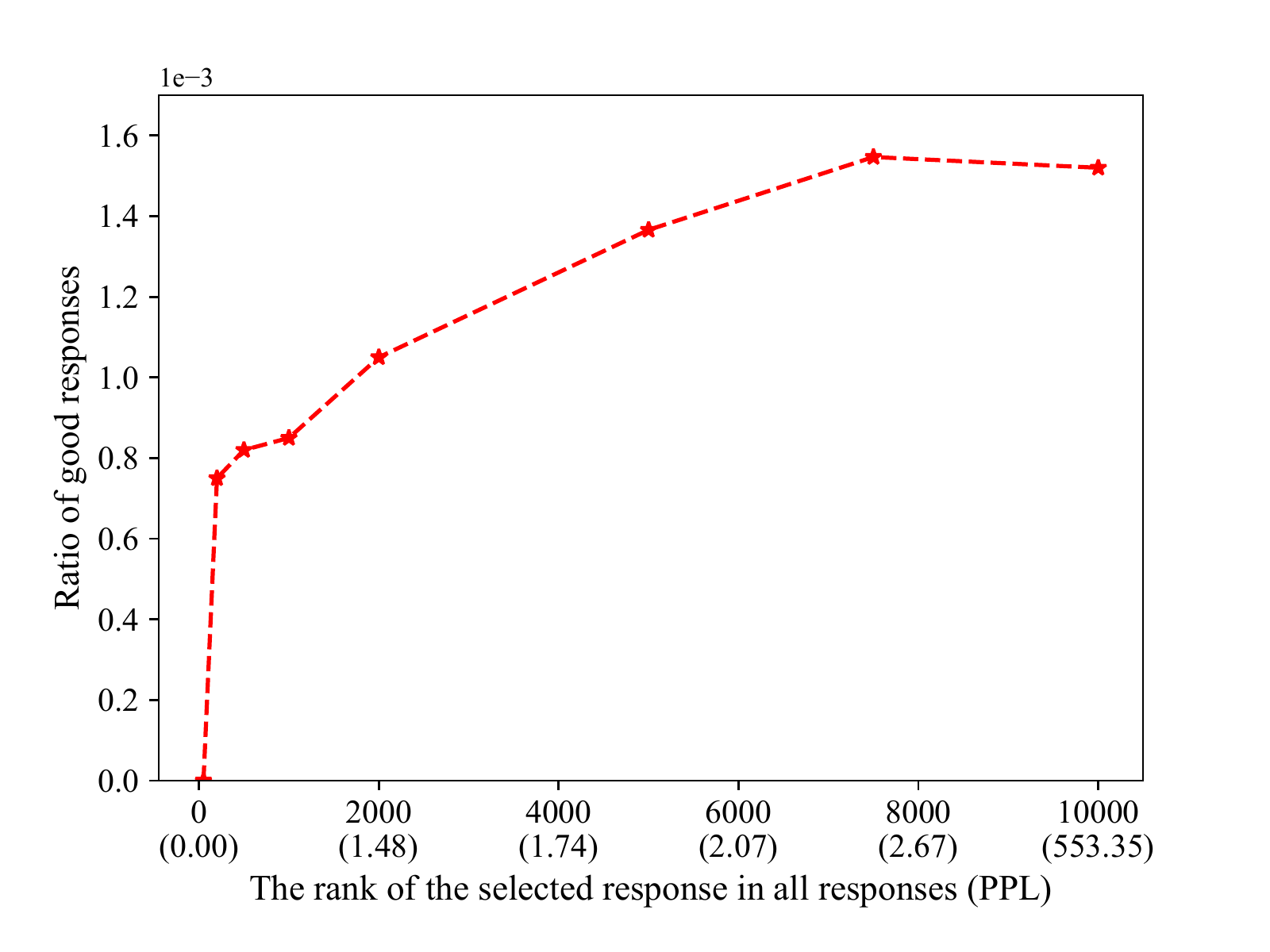}
    \caption{
    We use a dialogue model to generate 10,000 responses each for 100 utterances in PersonaChat and use perplexity (PPL) to rerank them.
   The response is good when the TF-IDF similarity between the response and ground truth is above 0.25 and the result of the natural language inference model between the response and persona information is entailment. PPL in brackets is the average value of all responses in each rank.}
    \label{fig:fig2}
\end{figure}

Open-domain dialogue systems need to give appropriate responses based on history utterances. An ideal open-domain dialogue system should generate consistent, coherent, and diverse responses. Part of the existing open-domain dialogue generation work focuses on improving the diversity of responses \citep{wang-etal-2021-diversifying}, while avoiding generating generic responses and achieving good results. How to improve the consistency of dialogue generation is also an urgent problem to be solved \cite{kim-etal-2020-will}. In addition, there is still the problem of poor coherence in dialogue generation \cite{ye-etal-2021-towards-quantifiable}.

To improve the consistency and coherence of dialogue generation, the existing works mainly improve from three aspects: valuable data construction and filtering \citep{zhang-etal-2018-personalizing, song-etal-2020-profile}, model structure modifying \cite{song-etal-2021-bob, zou-etal-2021-thinking, 2023DialoguePersona} and objective function designing \citep{li-etal-2020-dont, hao2020ranking}. However, the problem of poor consistency and coherence is still a tough challenge, especially in persona-based dialogue generation \cite{song-etal-2020-generate}. 
Because multiple constraints need to be satisfied simultaneously, part of which cannot be directly optimized, and part of the constraints conflict with each other, such as the conflict between the fluency of responses and the consistency of persona information.
In addition, the above methods need to retrain the model and can only adapt to the part of the existing dialogue models. For example, \citet{boyd-etal-2020-large} carefully design the objective function and scale model sizes from 117M to 8.3B parameters, which brings a lot of training costs. Fortunately, we find that the existing dialogue models actually have strong capabilities that can generate consistent and coherent responses, and we just need to find ways to release their capabilities.

First, we take a deep look at the characteristics of dialogue models, which believe that the response with the highest probability is the best. However, we wonder whether the high-probability responses generated by dialogue models are necessarily better than the low-probability responses. 
Based on the statistics in Figure~\ref{fig:fig2}, when the generation probability of responses decreases, the ratio of good responses increases first and then decreases. It shows that the ratio of good responses among low-probability responses is higher than that of high-probability responses, which is counter-intuitive. This is most likely because dialogue models use PPL as an optimization goal, but it is inconsistent with the requirements of coherence and consistency. To verify whether the good response with high TF-IDF similarity and high probability of entailment\footnote{The high TF-IDF similarity means the TF-IDF similarity between the response and ground truth is above 0.25, and the high probability of entailment means the entailment probability of the natural language inference model between the response and persona information is above 0.5. When the above two constraints are relaxed to 0.15 and 0.35, respectively, the trend of the curve in Figure~\ref{fig:fig2} is still the same.} is indeed superior to the response directly generated by the model, we use the human evaluation for experimental validation. As shown in Table~\ref{table0}, such responses are better than those directly generated by the model. Therefore, it only needs to sample large-scale diverse responses from existing dialogue models and then select good responses. 


Inspired by the aforementioned motivations, We propose a \textbf{sim}ple two-stage method: \textbf{o}ver-sampling \textbf{a}nd \textbf{p}ost-evaluation (SimOAP) to improve the coherence and consistency in persona-based dialogue. There are two challenges in our work. The large-scale sampling will bring additional time cost, how to accelerate the model is a challenge. Large-scale sampling can produce good responses, how to pick good responses from them is another challenge. We address the above two challenges using oversampling and post-evaluation, respectively.
In the over-sampling stage, SimOAP uses existing dialogue models for large-scale sampling, and the distilled or compressed models are used to reduce the additional time cost. In the post-evaluation stage, the TF-IDF algorithm \cite{salton1988term} and natural language inference (NLI) are used for coherence and consistency evaluation, respectively.


\begin{table}
\centering
\begin{tabular}{lcccc}
\hline
\textbf{} & \textbf{Flue} $\uparrow$ & \textbf{Cohe} $\uparrow$ & \textbf{Info} $\uparrow$ & \textbf{Cons} $\uparrow$ \\
\hline
DIR & 2.60 & 2.58 & 2.56 & 0.20 \\
S\&F & \textbf{3.40} & \textbf{3.36} & \textbf{3.42} & \textbf{0.72} \\
\hline
\end{tabular}
\caption{\label{table0}
Human evaluation results on responses generated by sampling and filtering (S\&F) or directly generated (DIR) from the dialogue model. We randomly select 50 examples from each of the above two. We evaluate the quality of the responses from fluency (\textbf{Flue}), coherence (\textbf{Cohe}), informativeness (\textbf{Info}) and consistency (\textbf{Cons}). Fluency, coherence and informativeness are scored on a scale of 1 to 5, consistency is 0 or 1.
}
\end{table}

To verify the effectiveness of our method, we conduct experiments on a persona-based dialogue dataset Personachat \citep{zhang-etal-2018-personalizing}. Automatic evaluations and human evaluations show that our method effectively boosts the performance of dialogue models and outperforms two baselines \cite{li-etal-2016-diversity,adiwardana2020towards} on most metrics. In addition, applying our method to small models can also achieve a better performance than using large models directly.

Our contributions in this paper are three folds:
\begin{itemize}
    \item We verify that the high-probability responses generated by dialogue models are not necessarily better than the low-probability responses. That is, dialogue models can generate good responses, but they are not selected.
    \item We propose a simple two-stage method: over-sampling and post-evaluation to improve the coherence and consistency in persona-based dialogue generation and it is model-agnostic.
    \item We conduct comprehensive experiments on a persona-based dialogue dataset. Automatic evaluations and human evaluations show that our method improves the backbone models and outperforms the baselines.
\end{itemize}


\section{Related Work}

\begin{figure*}[htp]
    \centering
    \includegraphics[width=16cm]{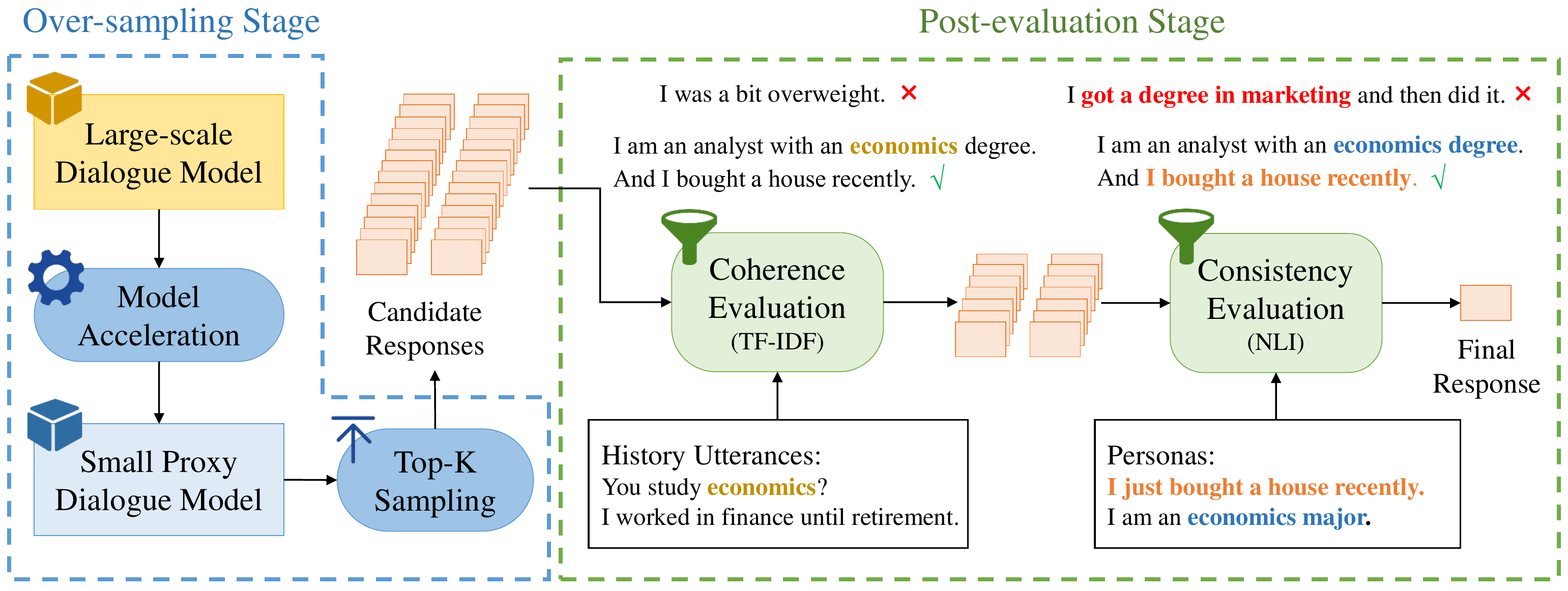}
    \caption{The framework of the proposed SimOAP method, which consists of an over-sampling stage and a post-evaluation stage. The post-evaluation stage consists of two parts: coherence evaluation and consistency evaluation. The text marked in the same color represents coherent or consistent text.}
    \label{fig:fig3}
\end{figure*}

Dialogue generation has made remarkable progress in recent years. Many pre-trained dialogue models have been proposed \citep{zhang2019dialogpt, bao2019plato, adiwardana2020towards, roller-etal-2021-recipes}.
To improve the consistency of dialogue generation and make dialogue models applicable to various scenarios, \citet{zhang-etal-2018-personalizing} propose a persona-based dialogue dataset PersonaChat. Persona-based dialogue generation is limited by the scale of data and expensive annotation costs. 
\citet{DBLP:journals/corr/abs-1911-05889} generate persona-based dialogue by using additional natural language inference data. \citet{cao-etal-2022-model} use data augmentation to extend data and use data distillation to make it easier to fit. However, labeling data for persona-based dialogue takes a high cost, and data from other domains is difficult to apply to persona-based dialogue fully.

Part of the work modifies the model structure for persona-based dialogue. \citet{DBLP:journals/corr/abs-1911-04700} propose a pre-trained model, which uses persona-based sparse data for pre-training. \citet{song-etal-2020-generate} design a three-stage framework of generating, deleting, and rewriting. 
\citet{song-etal-2021-bob} learn the persona features by designing a response generation decoder and a consistency understanding decoder. However, there are multiple constraints that need to be satisfied simultaneously, some of which cannot be directly optimized. The above works also bring a huge training cost.

Part of the work designs the related objective function. \citet{li-etal-2020-dont} modify the unlikelihood loss to improve the consistency of dialogue. \citet{boyd-etal-2020-large} use the previous dialogue content of users 
to control the dialogue of specific personas. However, it is difficult to design the objective function. We found a simple strategy without filtering valuable data, modifying the model structure, or designing objective functions, but only needs to use existing models for large-scale sampling and post-evaluation to improve the performance.

\citet{nye2021improving} use dual systems to improve the coherence and consistency of neural sequence models. This work uses a neural system to generate the candidate and a logic system to evaluate it. The candidate is generated and evaluated one by one until it meets the criteria. However, the small number of candidates limits the effectiveness of dialogue generation. In addition, the logic system evaluates the candidate by tracking common sense information. It is difficult to apply to dialogue generation. 
In dialogue generation, maximum mutual information (MMI) \citep{li-etal-2016-diversity} uses the mutual information between history utterances and responses to evaluate responses. MMI can reduce the generation of generic responses but brings the large-scale time cost.
To eliminate the influence of response length on likelihood, \citet{adiwardana2020towards} use length-normalized loglikelihood score (LLS) to evaluate candidate responses. However, it is verified that using large-scale sampling
for LLS performs worse than fewer candidate responses. It shows that LLS cannot release the ability of models by over-sampling. Furthermore, simple evaluation methods for the above two methods are difficult to work well in complex persona-based dialogue.

\section{Our Approach}

Persona-based dialogue consists of persona information sentences $P=\{p_{1},p_{2},...,p_{|P|}\}$, history utterances $H=\{h_{1},h_{2},...,h_{|H|}\}$, and a gold response $g$. Dialogue models need to generate a response $r$, which is coherent with history utterances $H$ and consistent with persona sentences $P$. 

The framework of SimOAP is shown in Figure~\ref{fig:fig3}. 
SimOAP consists of two stages: over-sampling and post-evaluation. In the over-sampling stage, SimOAP uses existing dialogue models for large-scale sampling, and accelerates the model to reduce the extra time cost. In the post-evaluation stage, the TF-IDF algorithm \cite{salton1988term} and natural language inference are used for coherence and consistency evaluation, respectively.

\subsection{Over-sampling Stage}
To do efficient and diverse over-sampling, we face two challenges to be solved. The first challenge is that generating large-scale responses is time-consuming, which will bring additional time cost. We have to speed it up. Another challenge is how to achieve diversity among different responses. The generated responses need to be diverse, not just those with high generation probability. Because we need to select a good response from the sampled responses, there should be differences between them rather than a large number of similar responses. To address the above challenges, we use distilled or compressed models to accelerate. Then the top-$k$ sampling \cite{fan2018hierarchical} with large $k$ value and large sample number are used to introduce diversity. The capabilities of well-trained dialogue models can be released by introducing diversity and large-scale sampling.

\paragraph{Generation of Candidate Responses}
The existing dialogue models actually have strong capabilities that can generate consistent and coherent responses, but they are just not being released. We choose existing dialogue models for dialogue generation without re-training.
To introduce diversity, we use top-$k$ sampling to take large-scale samples from existing dialogue models and generate candidate responses. 
At each step of generating the response, the dialogue model generates the probability of each word in the vocabulary being the likely next word, forming a probability distribution. Then we randomly sample from the $k$ most likely vocabs from this probability distribution. All tokens in each response are generated with top-$k$ sampling. 
To ensure the diversity of candidate responses and the effectiveness of over-sampling, we use the large $k$ in top-$k$ sampling. For each history dialogue, $s$ candidate responses will be generated, denoting them as $R=\{r_{1},r_{2},...,r_{s}\}$, and $s$ is also set to be large to introduce diversity.


\paragraph{Model Acceleration}
Due to the extra time cost incurred in large-scale sampling, we use distilled or compressed models to replace the backbone models to speed up. 
For example, when the backbone model is Multi-GPT2 \citep{cao-etal-2020-pretrained}, we use DistilGPT2 \citep{sanh2019distilbert} replace GPT2 \cite{radford2019language} to build Multi-GPT2. 
When the backbone model is BERT-over-BERT \citep{song-etal-2021-bob}, we use the compressed model BERT-medium \citep{devlin2018bert} replace BERT-base \citep{devlin2018bert} to build it. 

\subsection{Post-evaluation Stage}
The over-sampling stage produces diverse responses, but how to select good responses from them is a challenge. Although there are many metrics to automatically evaluate the effectiveness of dialogue \cite{gao-etal-2021-ream,chan-etal-2021-enhancing,ji-etal-2022-achieving,ghazarian-etal-2022-deam}, most of them evaluate the responses only from a single aspect. For example, perplexity can only be used to evaluate the fluency of responses and cannot reflect the quality of responses in other aspects. When multiple methods are used in combination to evaluate responses, it may bring additional time cost, especially for learnable methods. The oversampling stage already brings the additional time cost, so we want to reduce the time cost in the post-evaluation stage. How to reduce it is another challenge. 
To address the above challenges, we first use the TF-IDF algorithm to evaluate the coherence of candidate responses and filter out those with poor coherence\footnote{
Evaluating coherence using the TF-IDF algorithm is sufficient for our method to perform well and it is fast, which is verified in Section~\ref{sec:expen}.}.
Then the consistency evaluation with the NLI model is used to select the final response. Since both coherence and consistency need to be satisfied, the fast coherence evaluation based on TF-IDF is first used to evaluate and reduce candidate responses, which can reduce the time cost, then the learnable NLI is used.

\paragraph{Coherence Evaluation}
Coherence requires the response to be context-related to history utterances \cite{ye-etal-2021-towards-quantifiable}. Some learnable coherence evaluation methods \cite{ghazarian2022deam,ye-etal-2021-towards-quantifiable} have been proposed, but they will bring the additional time cost. 
To reduce the time cost of the post-evaluation stage and improve the coherence of responses, we use the TF-IDF algorithm \cite{salton1988term} to calculate the semantic similarity between the candidate responses $R$ and history utterances $H$. We take history utterances $H$ as the first document and each candidate response as a document, which together with $H$ constitute the corpus. The TF-IDF value of the $i$-th word $t_{i}$ in the corpus of the $j$-th document $d_{j}$ is: 
\begin{equation}\label{eq1}
\mathrm{tfidf}_{i,j}=\mathrm{tf}_{i,j}\ast \mathrm{idf}_{i},
\end{equation}
where $\mathrm{tf}_{i,j}$ is the term frequency of the $i$-th word in the $j$-th document, $\mathrm{idf}_{i}$ is the inverse document frequency of the $i$-th document:
\begin{equation}\label{eq2}
\mathrm{tf}_{i,j}=\frac{n_{i,j}}{\sum_{k} n_{k,j}}, 
\mathrm{idf}_{i}=\lg \frac{|D|}{1+\{j:t_{i}\in d_{j}\}},
\end{equation}
where $n_{i,j}$ is the number of the $i$-th word that appears in the $j$-th document, ${\sum_{k} n_{k,j}}$ is the sum of the number of all words in the $j$-th document. $|D|$ is the number of documents in the corpus, $\{j:t_{i}\in d_{j}\}$ is the number of documents which containing the $i$-th word $t_{i}$. Suppose there are $I$ unique words in the corpus, so each document vector can be represented as:
\begin{equation}\label{eq4}
V_{j}=\left (\mathrm{tfidf}_{1,j},...,\mathrm{tfidf}_{i,j}...,\mathrm{tfidf}_{I,j}  \right).
\end{equation}
Finally, we calculate the cosine similarity between the representation of $H$ and the representation of each candidate response $r_{i}$ separately, and $c$ responses with the highest similarity are selected as candidate $\hat{R}$, which is a subset of $R$.

\paragraph{Consistent Evaluation}
In persona-based dialogue, consistency requires the response to be consistent with persona information \cite{song-etal-2020-generate}. 
\citet{welleck-etal-2019-dialogue} propose a dialogue inference dataset DialogueNLI. Many persona-based dialogue works using NLI models fine-tuned on it to evaluate the consistency between persona information and responses have proven successful \cite{kim-etal-2020-will,DBLP:journals/corr/abs-1911-05889,song-etal-2020-generate,cao-etal-2022-model}.
Following them, we use the NLI model to calculate the possibility of entailment between the candidate responses $\hat{R}$ and persona sentences $P$ to improve the consistency. The RoBERTa \citep{DBLP:journals/corr/abs-1907-11692} is fine-tuned on DialogueNLI, where the inference labels are entailment, contradiction, or neutral. Then the fine-tuned RoBERTa is used to compute the possibility of entailment between candidate responses and persona sentences. Finally, the response with the highest possibility is selected as the final response $r$.

\label{sec:length}

\section{Experiments}
\label{sec:expen}
\subsection{Dataset}
To verify the effectiveness of our proposed method, experiments have been carried out on a public dialogue dataset \textbf{PersonaChat} \citep{zhang-etal-2018-personalizing}. PersonaChat is a persona-based dialogue dataset that includes rich persona features. During the dialogue process, the dialogue agent needs to give an appropriate response according to persona features. PersonaChat contains 10,907 dialogues (162,064 utterances), 8,939/1,000/968 dialogues of which are set for training/validation/testing.

\subsection{Experiment Settings}
\paragraph{Models and Baselines}
Two persona-based dialogue models and two baseline strategies are used for experimental verification.

\textbf{BERT-over-BERT (BoB)} \citep{song-etal-2021-bob} is a persona-based dialogue model which learns the persona features by designing an encoder, a response generation decoder, and a consistency understanding decoder.

\textbf{Multi-GPT2} \citep{cao-etal-2020-pretrained} is a persona-based dialogue model with encoder-decoder architecture adapted from GPT2. 

\textbf{Maximum Mutual Information (MMI)} \citep{li-etal-2016-diversity} use the backward model to predict history utterances from candidate responses. Then the prediction probability is used to rerank the responses and reduce generic responses.

\textbf{Length-normalized Loglikelihood Score (LLS)} \citep{adiwardana2020towards} is used to eliminate the influence of response length on likelihood. It is calculated as $\frac{\log_{}{P}}{T}$, where $P$ is the likelihood of the response and $T$ is the token number of the response. 

\begin{table*}[htp]
\centering
\scalebox{0.95}{
\begin{tabular}{lcccccc|cc}
\hline
\textbf{} & \textbf{PPL}$_\mathrm{BERT}$ $\downarrow$ & \textbf{PPL}$_\mathrm{GPT2}$ $\downarrow$ & \textbf{Dis-1} $\uparrow$ & \textbf{Dis-2} $\uparrow$ & \textbf{C} $\uparrow$ & \textbf{Avg} $\uparrow$ & \textbf{Rep} $\downarrow$ & \textbf{Avg-R} $\uparrow$\\
\hline
BoB & 42.47 & 139.04 & 5.62 & 17.77 & 0.114 & 0.262 & 8.63 & 0.326 \\
~ + MMI & 21.74 &  108.04 & 5.27 & 20.22 & 0.353 & 0.680 & 3.55 &  0.712 \\
~ + LLS & 19.34 &  81.96 & 5.20 & 17.21 & 0.048 & 0.444 & 23.10 & 0.370 \\
\hline
~ + SimOAP & \textbf{9.93} & \textbf{68.43} & 4.21 & 18.78 & \textbf{0.579}   & \textbf{0.704} & \textbf{0.65} & \textbf{0.754} \\
\hline
\hline
Multi-GPT2 & 109.76 & 361.40 & 3.92 & 29.57 & 0.145 & 0.542 & 1.65 & 0.612\\
~ + MMI & 281.99 & 1198.96 & 6.85 & 33.16 & 0.610 & 0.537 & 4.57 & 0.593\\
~ + LLS & \textbf{17.36} & \textbf{131.70} & 1.88 & 11.24 & 0.124 & 0.400 & 34.80 & 0.333\\
\hline
~ + SimOAP & 50.90 & 210.82 & 2.05 & 18.41 & \textbf{0.836} & 0.655 & 1.30 & 0.712\\
~ + SimOAP-Q & 58.76 & 244.62 & 2.38 & 20.95 & 0.814 & \textbf{0.671} & \textbf{0.93} & \textbf{0.724}\\
\hline
\end{tabular}
}
\caption{\label{table1}
Automatic indicators on PersonaChat dataset. Avg is the average of the min-max normalized score of each indicator except Rep. Avg-R is the average of the min-max normalized score of all indicators including Rep.}
\end{table*}

\paragraph{Implementation Details}
In the over-sampling stage, $k$ in top-$k$ sampling and the number of oversampling $s$ are set to 100 and 2000. After the coherence evaluation, 100 candidate responses with the highest similarity are retained. BoB has two decoders, the first decoder is used to generate a preliminary response and the second decoder is used to modify the preliminary response and generate the final response. We only use top-$k$ sampling in the first decoder. The second decoder is a response modifier, so we use greedy search. For Multi-GPT2, we directly use top-$k$ sampling for sampling. 
We keep the same as BoB\footnote{\url{https://github.com/songhaoyu/BoB}} and Multi-GPT2\footnote{\url{https://github.com/caoyu-noob/Multi-GPT2}} for the parameter settings of the model. For MMI, following as \citet{zhang2019dialogpt}, we use a pre-trained backward model DialoGPT-reverse to predict source utterances from candidate responses. Source utterances are composed of the concatenation of persona sentences and history utterances. The candidate responses are the same as our method. For LLS, we use the best parameters in \citet{adiwardana2020towards}: top-$k$ sampling is used to generate responses, $k$ is set to 40, and the number of responses generated is set to 20. The RoBERTa used in the consistency evaluation is RoBERTa-large. The experiments were completed via PyTorch on 4 32GB NVIDIA V100 GPUs.

\subsection{Evaluation Metrics}
\paragraph{Automatic Metrics}
In automatic evaluation, we choose the metrics in different aspects to evaluate the quality of responses. 
For diversity assessment, we use distinct-1/2 (\textbf{Dis-1/2}) \citep{li-etal-2016-diversity}. Furthermore, we propose a sentence-level repetition rate (\textbf{Rep}) for evaluating diversity. It is calculated as $Rep=\frac{n_{rep} }{N}$,
where $n_{rep}$ is the number of the responses which are the same as at least one other response and that response differs from the ground truth, $N$ is the total number of responses. 

For fluency assessment, we use perplexity (\textbf{PPL}) to evaluate the fluency of responses. GPT2 and BERT are chosen as language models to calculate the PPL of responses \cite{dathathri2019plug, qian2022controllable}, and calculation details are given in Appendix~\ref{sec:appendix_A}. For consistency assessment, we use consistency score (\textbf{C}) \citep{madotto-etal-2019-personalizing}. The BERT-large \citep{devlin2018bert} fine-tuned on DialogueNLI dataset \citep{welleck-etal-2019-dialogue} as NLI model is used to evaluate the consistency between persona sentences and responses. When the relation between them is entailment, neutral, and contradiction, C is 1, 0, and -1, respectively. To evaluate the overall performance of responses, we calculate the average of the min-max normalized score of each indicator except Rep, recorded as the average score (\textbf{Avg}). PPL and Rep are the lower the better, so use their negative numbers when calculating. The average score which includes Rep is recorded as \textbf{Avg-R}.

\begin{table*}[htp]
\centering
\scalebox{0.9}{
\begin{tabular}{lccccccc}
\hline
\textbf{} & \textbf{PPL}$_\mathrm{BERT}$ $\downarrow$ & \textbf{PPL}$_\mathrm{GPT2}$ $\downarrow$ & \textbf{Dis-1} $\uparrow$ & \textbf{Dis-2} $\uparrow$ & \textbf{C} $\uparrow$ & \textbf{Rep} $\downarrow$ & \textbf{Model-Size}\\
\hline
BoB$_\mathrm{base}$ & 42.47 & 139.04 & 5.62 & 17.77 & 0.114 & 8.63 & \multirow{2}*{1470MB} \\
~ + SimOAP & \textbf{9.93} & \textbf{68.43} & 4.21 & 18.78 & 0.579 & \textbf{0.65} &  \\
\hline
BoB$_\mathrm{medium}$ + SimOAP & 23.07 & 102.73 & \textbf{5.66} & \textbf{30.50} & \textbf{0.702} & 1.24 & 538MB \\
BoB$_\mathrm{mini}$ + SimOAP & 45.95 & 171.89 & 5.03 & 29.48 & 0.679 & 1.47 & 136MB \\
\hline
\hline
Multi-GPT2 & 109.76 & 361.40 & \textbf{3.92} & \textbf{29.57} & 0.145 & 1.65 & \multirow{2}*{1358MB} \\
~ + SimOAP & \textbf{58.76} & \textbf{244.62} & 2.38 & 20.95 & \textbf{0.836} & 0.93 \\
\hline
Multi-GPT2$_\mathrm{distil}$ + SimOAP & 66.41 & 247.27 & 2.46 & 21.13 & 0.823 & \textbf{0.56} & 829MB\\
\hline
\end{tabular}
}
\caption{\label{table2}
Automatic evaluation results of small models on PersonaChat dataset.}
\end{table*}

\begin{table}[htp]
\centering
\scalebox{0.77}{
\begin{tabular}{lcccc}
\hline
\textbf{} & \textbf{Flue} & \textbf{Cohe} & \textbf{Info} & \textbf{Cons} \\
\hline
BoB$_\mathrm{base}$ & 2.70 & 2.61 & 2.65 & 0.22 \\
~ + MMI & 3.02 & 3.07 & 3.02 & 0.49 \\
~ + LLS & 2.99 & 2.74 & 2.61 & 0.27 \\
~ + SimOAP & \textbf{3.59} & \textbf{3.43} & \textbf{3.55} & \textbf{0.70} \\
\hline
BoB$_\mathrm{medium}$ + SimOAP & 3.22 & 3.33 & 3.35 & \textbf{0.70} \\
\hline
\hline
Multi-GPT2$_\mathrm{base}$ & 2.64 & 2.41 & 2.62 & 0.17 \\
~ + MMI & 3.04 & 3.01 & 3.02 & 0.57 \\
~ + LLS & 3.05 & 2.85 & 2.36 & 0.26 \\
~ + SimOAP & 3.14 & 3.06 & 2.85 & 0.68  \\
~ + SimOAP-Q & \textbf{3.38} & \textbf{3.33} & 3.33 & 0.57  \\
\hline
Multi-GPT2$_\mathrm{distil}$ + SimOAP & 3.13 & 3.22 & \textbf{3.47} & \textbf{0.72} \\
\hline
\end{tabular}
}
\caption{\label{table3}
Human evaluation results on PersonaChat.
}
\end{table}


\paragraph{Human Evaluations}
We randomly select 50 examples each from the baselines and our method for human evaluation. Three graduate students with good English skills are asked to rate the quality of responses from fluency (\textbf{Flue}), coherence (\textbf{Cohe}), informativeness (\textbf{Info}), and consistency (\textbf{Cons}). Fluency, coherence, and informativeness are scored on a scale of 1 to 5, where 5 is good, 3 is moderate, and 1 is poor. The score for consistency is 0 or 1, where 0 indicates that the response is inconsistent or irrelevant to persona sentences, and 1 indicates that the response is relevant and consistent with persona sentences.

\subsection{Results}
\paragraph{Results on Full-size Models}
As shown in Table~\ref{table1}, our method surpasses two backbone models on all automatic metrics except Dis-1/2, indicating that our method can effectively improve the performance of persona-based dialogue models.

Our method outperforms MMI on all automatic metrics except Dis-1/2, indicating that our post-evaluation stage can select the better response from candidate responses. Furthermore, the generation speed of our method is faster than MMI\footnote{The details of experiments with generation speed are given in Appendix~\ref{sec:appendix}.}. 

For LLS, the responses generated by our method outperform it in almost all metrics. Only responses generated by Multi-GPT2 using LLS are lower than those generated by our method on PPL. However, the responses generated by Multi-GPT2 using the LLS have many repetitive responses, of which Rep is 34.80\%. The Rep of our method is only 0.93\%, indicating that the over-sampling stage of our method can effectively generate diverse responses. 
Although LLS is faster than our method for generation speeds, it is on average 0.33 lower than our method on two average scores. It is also significantly lower than MMI. In addition, the overall performance of our method outperforms all backbone models and baselines on Avg and Avg-R.

Finally, we use human evaluation to further evaluate responses. As shown in Table~\ref{table3}, our method outperforms backbone models and baselines on all metrics. It shows that the responses generated by our method are more fluent and informative. Meanwhile, they are more coherent to history utterances and more consistent with persona sentences.

\paragraph{Further Analysis of SimOAP}
First, we analyze the reasons for the choice of method in the post-evaluation stage. As shown in Table~\ref{table_A2} of Appendix~\ref{sec:appendix_C}, the time cost of the learnable coherence evaluation method approaches or even exceeds the generation time of Multi-GPT2, which is unacceptable. The TF-IDF algorithm is fast and shows a good evaluation effect, so we choose it.

Furthermore, we compare the effectiveness of Multi-GPT2 using all history dialogue and only the last two sentences of it in the coherence evaluation. The average score of the latter (Multi-GPT2 + SimOAP-Q in Table~\ref{table1}) is slightly higher. We think the reason is that too much history dialogue will cause interference. BoB only uses the last utterance of the history dialogue to generate responses, so we do not need to verify it.

\paragraph{Results on Accelerated Models}
To speed up our method and verify whether small models using SimOAP can surpass large models, we use BERT-medium and BERT-mini to replace the BERT-base used in BoB. As shown in Table~\ref{table2}, the BERT-medium-based BoB using our method outperforms BoB on PPL, and its size is only 36.6\% of BoB. It is worth noting that the BERT-medium-based BoB using SimOAP to generate responses significantly improves diversity and consistency.
The BERT-mini-based BoB performs worse than BoB on PPL, which indicates that the original ability of models is also important.
For Multi-GPT2, we use DistilGPT2 to replace the GPT2 used in it. After using our method, DistilGPT2-based Multi-GPT2 also surpasses Multi-GPT2 on PPL and consistency score, and its size is only 61.05\% of Multi-GPT2. 


\begin{table*}[htp]
\centering
\scalebox{1}{
\begin{tabular}{l|l}
\hline
\multirow{4}*{\textbf{Persona}} & 1. I also work as a custodian to help pay the bills.\\
 & 2. I play the piano and guitar and sing.\\
 & 3. My favorite type of music to sing is folk music. \\
 & 4. I'm a musician and hope to make it big some day.\\
\hline
\textbf{History Utterances} & That's interesting. What instruments do you play?\\
\hline
BoB$_\mathrm{base}$ & I'm a bit of a bit of a comedian.\\
~ + MMI & Yes, i play the guitar.... I was a slave for many years. \\
~ + LLS & I play the piano and juggler..... How do you feel? \\
~ + SimOAP & I play the piano... And I have a few different instruments.\\
\hline
BoB$_\mathrm{medium}$ + SimOAP & I play the guitar and that's what I play. \\
\hline
\end{tabular}
}
\caption{\label{table5}
Generated examples from our method and baselines.
}
\end{table*}

\begin{figure}[htp]
    \centering 
    \includegraphics[width=7.5cm]{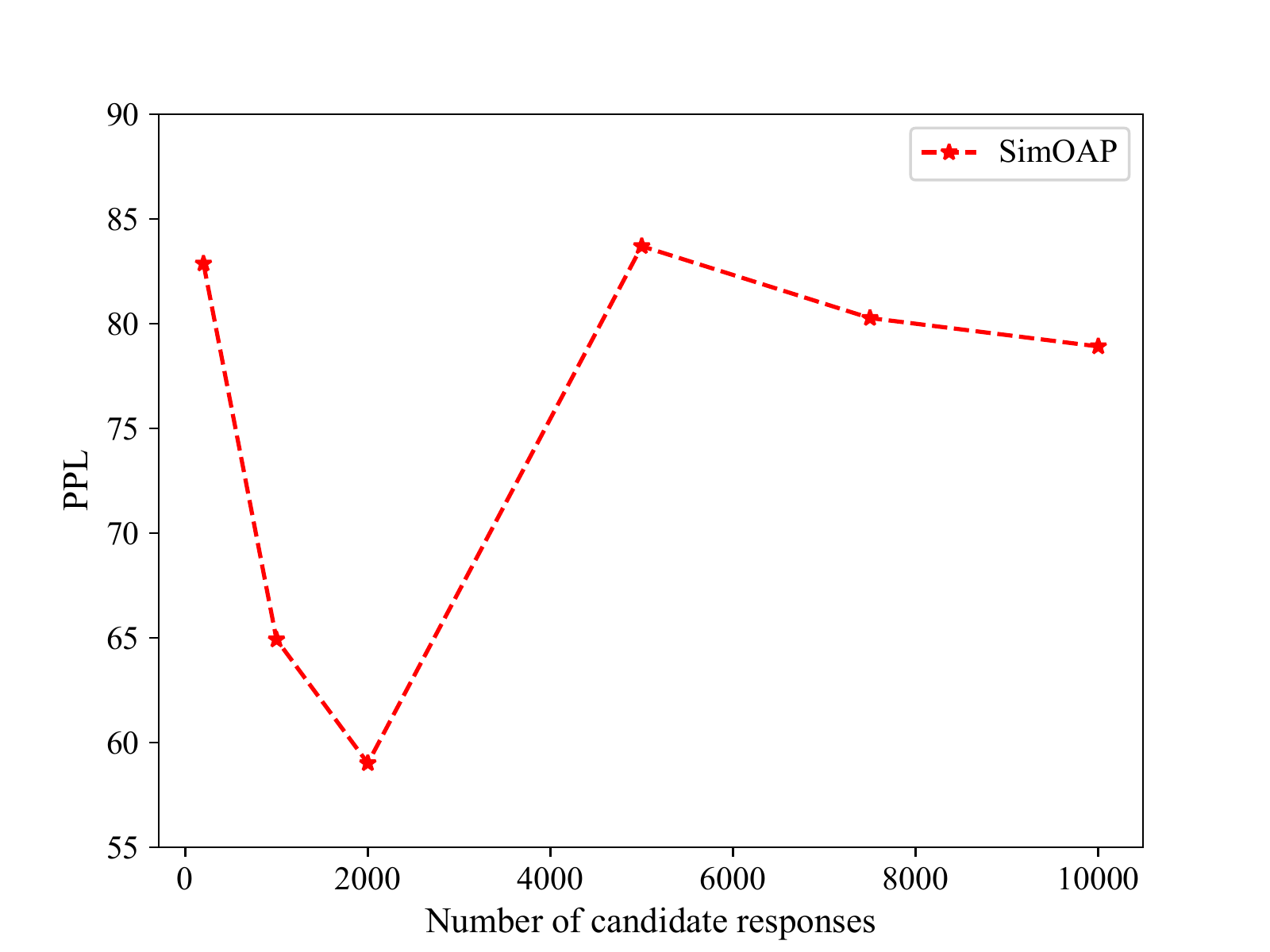}
    \caption{The impact of the number of candidate responses generated with our method on PPL. The PPL is calculated by GPT2.}
    \label{fig:fig4}
\end{figure}

\begin{figure}[htp]
    \centering 
    \includegraphics[width=7.5cm]{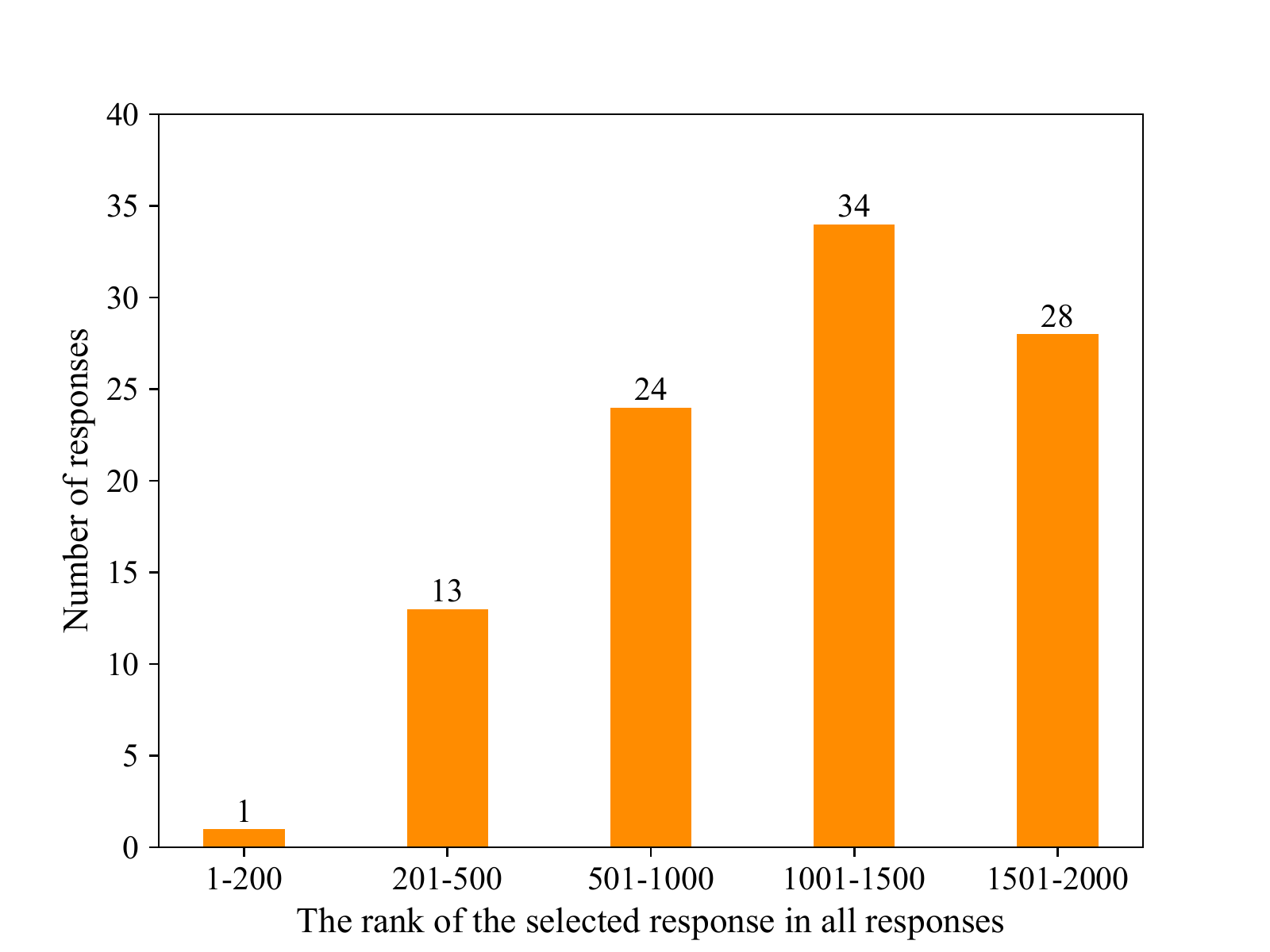}
    \caption{The number of responses selected by SimOAP in each probability generation interval.}
    \label{fig:fig6}
\end{figure}

\paragraph{Number of Candidate Responses Generated}
To verify the impact of generating different numbers of candidate responses on the performance of SimOAP, we use 100 pieces of data in PersonaChat for experimental verification. BoB is used to generate different numbers of candidate responses, and post-evaluation is used to select the final responses. We use PPL to evaluate the response, and PPL is computed by GPT2.
As shown in Figure~\ref{fig:fig4}, the PPL of generating 2000 candidate responses is lower than generating 200 or 1000 candidate responses.
The PPL increases when the number of candidate responses is further scaled up to 5000, 7500, or 10000. We believe that the plethora of candidate responses disrupts the post-evaluation stage. 
Thus, we set the number of generated candidate responses as 2000. In addition, we use the PPL of the backbone model to rerank the candidate responses.
The rank of the selected responses is calculated, and the results are shown in Figure~\ref{fig:fig6}. The average rank of the selected responses is 1135, and 86 responses are located in the rank with the PPL from 500th to 2000th. This shows that sometimes the dialogue model can generate good responses, but they are just not selected.

\begin{table}[htp]
\centering
\scalebox{0.77}{
\begin{tabular}{lcccc}
\hline
\textbf{} & \textbf{PPL}$_\mathrm{GPT2}$ & \textbf{Dis-1} & \textbf{Dis-2} & \textbf{C}\\ 
\hline
BoB + SimOAP & 68.43 & 4.21 & 18.78 & 0.579\\
\hline
~ $w/o$ TF-IDF & 79.28 & 4.31 & 18.44 & 0.818 \\ 
~ $w/o$ NLI & 105.84 & 5.97 & 23.00 & 0.070\\ 
\hline
Multi-GPT2 + SimOAP & 244.62 & 2.38 & 20.95 & 0.836\\
\hline
~ $w/o$ TF-IDF & 292.85 & 2.81 & 22.53 & 0.892\\ 
~ $w/o$ NLI  & 288.87 & 2.68 & 21.95 & 0.127\\
\hline
\end{tabular}
}
\caption{\label{citation-guide}
Ablation results of automatic metrics.
}
\end{table}

\subsection{Ablation Study}
To verify the effectiveness of coherence evaluation and consistency evaluation, we conduct ablation experiments. As shown in Table~\ref{citation-guide}, when only the coherence evaluation is used, the PPL of the responses increases, indicating that the fluency of the sentences has become worse. The consistency between the responses and the persona sentences also reduce. When only the consistency evaluation is used, although the consistency score is further improved, the PPL of the responses increases, which means the fluency of responses is reduced. Therefore, consistency evaluation and consistency evaluation in the SimOAP method are essential. Finally, we present an example generated using our method and baselines, as shown in Table~\ref{table5}.

\section{Conclusion}
In this work, we propose a simple but effective two-stage strategy to improve the coherence and consistency in persona-based dialogue generation. 
In the over-sampling stage, we use dialogue models for large-scale sampling, and compressed or distilled models are used to accelerate.
In the post-evaluation stage, multiple well-designed evaluation metrics select the final response from large-scale candidates.
Experimental results show that our method improves the backbone models and outperforms the baseline strategies.
For reproducibility, we publish the source code\footnote{\url{https://github.com/934865517zjk/SimOAP}}.
In future work, we will consider further acceleration of our method.

\section*{Limitations}
In this work, we generate diverse responses through large-scale sampling in the oversampling stage. Although we use the compression and distillation models to speed up, the problem of generation speed still exists. Thus, one of the limitations of this work is the additional time cost when generating large-scale candidate responses. In addition, we use existing dialogue models for dialogue generation, mainly used in short text generation and unsuitable for long text generation, which is another limitation of this work.

\section*{Ethics Statement}
Persona-based dialogue generation aims to improve the consistency of open-domain dialogue generation while enabling dialogue generation to be extended to more application scenarios. In persona-based dialogue, the dialogue model uses persona information in the process of dialogue generation. The purpose of using persona information is to improve the consistency of the dialogue system rather than guessing user identities or associating persona information with real-world users. In this work, we use public datasets and do not involve aggression or privacy concerns. Furthermore, we use existing dialogue models for research, so we have the same concerns as other dialogue generation research. For example, there is a risk of generating toxic or biased language.

\section*{Acknowledgements}
This work was supported by the National Key R\&D Program of China (2022YFB3103700, 2022YFB3103704), the National Natural Science Foundation of China (NSFC) under Grants No. 62276248, and the Youth Innovation Promotion Association CAS under Grants No. 2023111.

\bibliography{anthology,custom}

\begin{thebibliography}{37}
\expandafter\ifx\csname natexlab\endcsname\relax\def\natexlab#1{#1}\fi

\bibitem[{Adiwardana et~al.(2020)Adiwardana, Luong, So, Hall, Fiedel,
  Thoppilan, Yang, Kulshreshtha, Nemade, Lu et~al.}]{adiwardana2020towards}
Daniel Adiwardana, Minh-Thang Luong, David~R So, Jamie Hall, Noah Fiedel, Romal
  Thoppilan, Zi~Yang, Apoorv Kulshreshtha, Gaurav Nemade, Yifeng Lu, et~al.
  2020.
\newblock Towards a human-like open-domain chatbot.
\newblock \emph{arXiv preprint arXiv:2001.09977}.

\bibitem[{Bao et~al.(2019)Bao, He, Wang, Wu, and Wang}]{bao2019plato}
Siqi Bao, Huang He, Fan Wang, Hua Wu, and Haifeng Wang. 2019.
\newblock Plato: Pre-trained dialogue generation model with discrete latent
  variable.
\newblock \emph{arXiv preprint arXiv:1910.07931}.

\bibitem[{Boyd et~al.(2020)Boyd, Puri, Shoeybi, Patwary, and
  Catanzaro}]{boyd-etal-2020-large}
Alex Boyd, Raul Puri, Mohammad Shoeybi, Mostofa Patwary, and Bryan Catanzaro.
  2020.
\newblock \href {https://doi.org/10.18653/v1/2020.acl-main.8} {Large scale
  multi-actor generative dialog modeling}.
\newblock In \emph{Proceedings of the 58th Annual Meeting of the Association
  for Computational Linguistics}, pages 66--84, Online. Association for
  Computational Linguistics.

\bibitem[{Cao et~al.(2022)Cao, Bi, Fang, Shi, and Tao}]{cao-etal-2022-model}
Yu~Cao, Wei Bi, Meng Fang, Shuming Shi, and Dacheng Tao. 2022.
\newblock \href {https://doi.org/10.18653/v1/2022.acl-long.550} {A
  model-agnostic data manipulation method for persona-based dialogue
  generation}.
\newblock In \emph{Proceedings of the 60th Annual Meeting of the Association
  for Computational Linguistics (Volume 1: Long Papers)}, pages 7984--8002,
  Dublin, Ireland. Association for Computational Linguistics.

\bibitem[{Cao et~al.(2020)Cao, Bi, Fang, and Tao}]{cao-etal-2020-pretrained}
Yu~Cao, Wei Bi, Meng Fang, and Dacheng Tao. 2020.
\newblock \href {https://doi.org/10.18653/v1/2020.findings-emnlp.81}
  {Pretrained language models for dialogue generation with multiple input
  sources}.
\newblock In \emph{Findings of the Association for Computational Linguistics:
  EMNLP 2020}, pages 909--917, Online. Association for Computational
  Linguistics.

\bibitem[{Chan et~al.(2021)Chan, Liu, Li, Zhang, Zhao, Shi, and
  Yan}]{chan-etal-2021-enhancing}
Zhangming Chan, Lemao Liu, Juntao Li, Haisong Zhang, Dongyan Zhao, Shuming Shi,
  and Rui Yan. 2021.
\newblock \href {https://doi.org/10.18653/v1/2021.findings-acl.432} {Enhancing
  the open-domain dialogue evaluation in latent space}.
\newblock In \emph{Findings of the Association for Computational Linguistics:
  ACL-IJCNLP 2021}, pages 4889--4900, Online. Association for Computational
  Linguistics.

\bibitem[{Dathathri et~al.(2019)Dathathri, Madotto, Lan, Hung, Frank, Molino,
  Yosinski, and Liu}]{dathathri2019plug}
Sumanth Dathathri, Andrea Madotto, Janice Lan, Jane Hung, Eric Frank, Piero
  Molino, Jason Yosinski, and Rosanne Liu. 2019.
\newblock Plug and play language models: A simple approach to controlled text
  generation.
\newblock \emph{arXiv preprint arXiv:1912.02164}.

\bibitem[{Devlin et~al.(2018)Devlin, Chang, Lee, and
  Toutanova}]{devlin2018bert}
Jacob Devlin, Ming-Wei Chang, Kenton Lee, and Kristina Toutanova. 2018.
\newblock Bert: Pre-training of deep bidirectional transformers for language
  understanding.
\newblock \emph{arXiv preprint arXiv:1810.04805}.

\bibitem[{Fan et~al.(2018)Fan, Lewis, and Dauphin}]{fan2018hierarchical}
Angela Fan, Mike Lewis, and Yann Dauphin. 2018.
\newblock Hierarchical neural story generation.
\newblock \emph{arXiv preprint arXiv:1805.04833}.

\bibitem[{Gao et~al.(2021)Gao, Bi, Xu, and Shi}]{gao-etal-2021-ream}
Jun Gao, Wei Bi, Ruifeng Xu, and Shuming Shi. 2021.
\newblock \href {https://doi.org/10.18653/v1/2021.findings-acl.220}
  {{REAM}$\sharp$: An enhancement approach to reference-based evaluation
  metrics for open-domain dialog generation}.
\newblock In \emph{Findings of the Association for Computational Linguistics:
  ACL-IJCNLP 2021}, pages 2487--2500, Online. Association for Computational
  Linguistics.

\bibitem[{Ghazarian et~al.(2022{\natexlab{a}})Ghazarian, Wen, Galstyan, and
  Peng}]{ghazarian-etal-2022-deam}
Sarik Ghazarian, Nuan Wen, Aram Galstyan, and Nanyun Peng. 2022{\natexlab{a}}.
\newblock \href {https://doi.org/10.18653/v1/2022.acl-long.57} {{DEAM}:
  Dialogue coherence evaluation using {AMR}-based semantic manipulations}.
\newblock In \emph{Proceedings of the 60th Annual Meeting of the Association
  for Computational Linguistics (Volume 1: Long Papers)}, pages 771--785,
  Dublin, Ireland. Association for Computational Linguistics.

\bibitem[{Ghazarian et~al.(2022{\natexlab{b}})Ghazarian, Wen, Galstyan, and
  Peng}]{ghazarian2022deam}
Sarik Ghazarian, Nuan Wen, Aram Galstyan, and Nanyun Peng. 2022{\natexlab{b}}.
\newblock Deam: Dialogue coherence evaluation using amr-based semantic
  manipulations.
\newblock \emph{arXiv preprint arXiv:2203.09711}.

\bibitem[{Hao et~al.(2020)Hao, Pang, Lan, Sun, Guo, and Cheng}]{hao2020ranking}
Changying Hao, Liang Pang, Yanyan Lan, Fei Sun, Jiafeng Guo, and Xueqi Cheng.
  2020.
\newblock Ranking enhanced dialogue generation.
\newblock In \emph{Proceedings of the 29th ACM International Conference on
  Information \& Knowledge Management}, pages 465--474.

\bibitem[{Ji et~al.(2022)Ji, Graham, Jones, Lyu, and
  Liu}]{ji-etal-2022-achieving}
Tianbo Ji, Yvette Graham, Gareth Jones, Chenyang Lyu, and Qun Liu. 2022.
\newblock \href {https://doi.org/10.18653/v1/2022.acl-long.445} {Achieving
  reliable human assessment of open-domain dialogue systems}.
\newblock In \emph{Proceedings of the 60th Annual Meeting of the Association
  for Computational Linguistics (Volume 1: Long Papers)}, pages 6416--6437,
  Dublin, Ireland. Association for Computational Linguistics.

\bibitem[{Kim et~al.(2020)Kim, Kim, and Kim}]{kim-etal-2020-will}
Hyunwoo Kim, Byeongchang Kim, and Gunhee Kim. 2020.
\newblock \href {https://doi.org/10.18653/v1/2020.emnlp-main.65} {Will {I}
  sound like me? improving persona consistency in dialogues through pragmatic
  self-consciousness}.
\newblock In \emph{Proceedings of the 2020 Conference on Empirical Methods in
  Natural Language Processing (EMNLP)}, pages 904--916, Online. Association for
  Computational Linguistics.

\bibitem[{Li et~al.(2016)Li, Galley, Brockett, Gao, and
  Dolan}]{li-etal-2016-diversity}
Jiwei Li, Michel Galley, Chris Brockett, Jianfeng Gao, and Bill Dolan. 2016.
\newblock \href {https://doi.org/10.18653/v1/N16-1014} {A diversity-promoting
  objective function for neural conversation models}.
\newblock In \emph{Proceedings of the 2016 Conference of the North {A}merican
  Chapter of the Association for Computational Linguistics: Human Language
  Technologies}, pages 110--119, San Diego, California. Association for
  Computational Linguistics.

\bibitem[{Li et~al.(2020)Li, Roller, Kulikov, Welleck, Boureau, Cho, and
  Weston}]{li-etal-2020-dont}
Margaret Li, Stephen Roller, Ilia Kulikov, Sean Welleck, Y-Lan Boureau,
  Kyunghyun Cho, and Jason Weston. 2020.
\newblock \href {https://doi.org/10.18653/v1/2020.acl-main.428} {Don{'}t say
  that! making inconsistent dialogue unlikely with unlikelihood training}.
\newblock In \emph{Proceedings of the 58th Annual Meeting of the Association
  for Computational Linguistics}, pages 4715--4728, Online. Association for
  Computational Linguistics.

\bibitem[{Liu et~al.(2023)Liu, Huang, Xiechi, Wang, de~Melo, Lin, Pang, and
  He}]{2023DialoguePersona}
Pingsheng Liu, Zhengjie Huang, Zhang Xiechi, Linlin Wang, Gerard de~Melo, Xin
  Lin, Liang Pang, and Liang He. 2023.
\newblock A disentangled-attention based framework with persona-aware prompt
  learning for dialogue generation.
\newblock In \emph{Proceedings of AAAI 2023}. AAAI.

\bibitem[{Liu et~al.(2019)Liu, Ott, Goyal, Du, Joshi, Chen, Levy, Lewis,
  Zettlemoyer, and Stoyanov}]{DBLP:journals/corr/abs-1907-11692}
Yinhan Liu, Myle Ott, Naman Goyal, Jingfei Du, Mandar Joshi, Danqi Chen, Omer
  Levy, Mike Lewis, Luke Zettlemoyer, and Veselin Stoyanov. 2019.
\newblock \href {http://arxiv.org/abs/1907.11692} {Roberta: {A} robustly
  optimized {BERT} pretraining approach}.
\newblock \emph{CoRR}, abs/1907.11692.

\bibitem[{Madotto et~al.(2019)Madotto, Lin, Wu, and
  Fung}]{madotto-etal-2019-personalizing}
Andrea Madotto, Zhaojiang Lin, Chien-Sheng Wu, and Pascale Fung. 2019.
\newblock \href {https://doi.org/10.18653/v1/P19-1542} {Personalizing dialogue
  agents via meta-learning}.
\newblock In \emph{Proceedings of the 57th Annual Meeting of the Association
  for Computational Linguistics}, pages 5454--5459, Florence, Italy.
  Association for Computational Linguistics.

\bibitem[{Nye et~al.(2021)Nye, Tessler, Tenenbaum, and Lake}]{nye2021improving}
Maxwell Nye, Michael Tessler, Josh Tenenbaum, and Brenden~M Lake. 2021.
\newblock Improving coherence and consistency in neural sequence models with
  dual-system, neuro-symbolic reasoning.
\newblock \emph{Advances in Neural Information Processing Systems},
  34:25192--25204.

\bibitem[{Qian et~al.(2022)Qian, Dong, Shen, Wei, and
  Chen}]{qian2022controllable}
Jing Qian, Li~Dong, Yelong Shen, Furu Wei, and Weizhu Chen. 2022.
\newblock Controllable natural language generation with contrastive prefixes.
\newblock \emph{arXiv preprint arXiv:2202.13257}.

\bibitem[{Radford et~al.(2019)Radford, Wu, Child, Luan, Amodei, Sutskever
  et~al.}]{radford2019language}
Alec Radford, Jeffrey Wu, Rewon Child, David Luan, Dario Amodei, Ilya
  Sutskever, et~al. 2019.
\newblock Language models are unsupervised multitask learners.
\newblock \emph{OpenAI blog}, 1(8):9.

\bibitem[{Roller et~al.(2021)Roller, Dinan, Goyal, Ju, Williamson, Liu, Xu,
  Ott, Smith, Boureau, and Weston}]{roller-etal-2021-recipes}
Stephen Roller, Emily Dinan, Naman Goyal, Da~Ju, Mary Williamson, Yinhan Liu,
  Jing Xu, Myle Ott, Eric~Michael Smith, Y-Lan Boureau, and Jason Weston. 2021.
\newblock \href {https://doi.org/10.18653/v1/2021.eacl-main.24} {Recipes for
  building an open-domain chatbot}.
\newblock In \emph{Proceedings of the 16th Conference of the European Chapter
  of the Association for Computational Linguistics: Main Volume}, pages
  300--325, Online. Association for Computational Linguistics.

\bibitem[{Salton and Buckley(1988)}]{salton1988term}
Gerard Salton and Christopher Buckley. 1988.
\newblock Term-weighting approaches in automatic text retrieval.
\newblock \emph{Information processing \& management}, 24(5):513--523.

\bibitem[{Sanh et~al.(2019)Sanh, Debut, Chaumond, and
  Wolf}]{sanh2019distilbert}
Victor Sanh, Lysandre Debut, Julien Chaumond, and Thomas Wolf. 2019.
\newblock Distilbert, a distilled version of bert: smaller, faster, cheaper and
  lighter.
\newblock \emph{arXiv preprint arXiv:1910.01108}.

\bibitem[{Song et~al.(2021)Song, Wang, Zhang, Zhang, and
  Liu}]{song-etal-2021-bob}
Haoyu Song, Yan Wang, Kaiyan Zhang, Wei-Nan Zhang, and Ting Liu. 2021.
\newblock \href {https://doi.org/10.18653/v1/2021.acl-long.14} {{B}o{B}: {BERT}
  over {BERT} for training persona-based dialogue models from limited
  personalized data}.
\newblock In \emph{Proceedings of the 59th Annual Meeting of the Association
  for Computational Linguistics and the 11th International Joint Conference on
  Natural Language Processing (Volume 1: Long Papers)}, pages 167--177, Online.
  Association for Computational Linguistics.

\bibitem[{Song et~al.(2020{\natexlab{a}})Song, Wang, Zhang, Liu, and
  Liu}]{song-etal-2020-generate}
Haoyu Song, Yan Wang, Wei-Nan Zhang, Xiaojiang Liu, and Ting Liu.
  2020{\natexlab{a}}.
\newblock \href {https://doi.org/10.18653/v1/2020.acl-main.516} {Generate,
  delete and rewrite: A three-stage framework for improving persona consistency
  of dialogue generation}.
\newblock In \emph{Proceedings of the 58th Annual Meeting of the Association
  for Computational Linguistics}, pages 5821--5831, Online. Association for
  Computational Linguistics.

\bibitem[{Song et~al.(2020{\natexlab{b}})Song, Wang, Zhang, Zhao, Liu, and
  Liu}]{song-etal-2020-profile}
Haoyu Song, Yan Wang, Wei-Nan Zhang, Zhengyu Zhao, Ting Liu, and Xiaojiang Liu.
  2020{\natexlab{b}}.
\newblock \href {https://doi.org/10.18653/v1/2020.emnlp-main.539} {Profile
  consistency identification for open-domain dialogue agents}.
\newblock In \emph{Proceedings of the 2020 Conference on Empirical Methods in
  Natural Language Processing (EMNLP)}, pages 6651--6662, Online. Association
  for Computational Linguistics.

\bibitem[{Song et~al.(2019)Song, Zhang, Hu, and
  Liu}]{DBLP:journals/corr/abs-1911-05889}
Haoyu Song, Wei{-}Nan Zhang, Jingwen Hu, and Ting Liu. 2019.
\newblock \href {http://arxiv.org/abs/1911.05889} {Generating persona
  consistent dialogues by exploiting natural language inference}.
\newblock \emph{CoRR}, abs/1911.05889.

\bibitem[{Wang et~al.(2021)Wang, Zheng, Jiang, and
  Huang}]{wang-etal-2021-diversifying}
Yida Wang, Yinhe Zheng, Yong Jiang, and Minlie Huang. 2021.
\newblock \href {https://doi.org/10.18653/v1/2021.acl-long.272} {Diversifying
  dialog generation via adaptive label smoothing}.
\newblock In \emph{Proceedings of the 59th Annual Meeting of the Association
  for Computational Linguistics and the 11th International Joint Conference on
  Natural Language Processing (Volume 1: Long Papers)}, pages 3507--3520,
  Online. Association for Computational Linguistics.

\bibitem[{Welleck et~al.(2019)Welleck, Weston, Szlam, and
  Cho}]{welleck-etal-2019-dialogue}
Sean Welleck, Jason Weston, Arthur Szlam, and Kyunghyun Cho. 2019.
\newblock \href {https://doi.org/10.18653/v1/P19-1363} {Dialogue natural
  language inference}.
\newblock In \emph{Proceedings of the 57th Annual Meeting of the Association
  for Computational Linguistics}, pages 3731--3741, Florence, Italy.
  Association for Computational Linguistics.

\bibitem[{Ye et~al.(2021)Ye, Lu, Huang, Lin, and
  Liang}]{ye-etal-2021-towards-quantifiable}
Zheng Ye, Liucun Lu, Lishan Huang, Liang Lin, and Xiaodan Liang. 2021.
\newblock \href {https://doi.org/10.18653/v1/2021.acl-long.211} {Towards
  quantifiable dialogue coherence evaluation}.
\newblock In \emph{Proceedings of the 59th Annual Meeting of the Association
  for Computational Linguistics and the 11th International Joint Conference on
  Natural Language Processing (Volume 1: Long Papers)}, pages 2718--2729,
  Online. Association for Computational Linguistics.

\bibitem[{Zhang et~al.(2018)Zhang, Dinan, Urbanek, Szlam, Kiela, and
  Weston}]{zhang-etal-2018-personalizing}
Saizheng Zhang, Emily Dinan, Jack Urbanek, Arthur Szlam, Douwe Kiela, and Jason
  Weston. 2018.
\newblock \href {https://doi.org/10.18653/v1/P18-1205} {Personalizing dialogue
  agents: {I} have a dog, do you have pets too?}
\newblock In \emph{Proceedings of the 56th Annual Meeting of the Association
  for Computational Linguistics (Volume 1: Long Papers)}, pages 2204--2213,
  Melbourne, Australia. Association for Computational Linguistics.

\bibitem[{Zhang et~al.(2019)Zhang, Sun, Galley, Chen, Brockett, Gao, Gao, Liu,
  and Dolan}]{zhang2019dialogpt}
Yizhe Zhang, Siqi Sun, Michel Galley, Yen-Chun Chen, Chris Brockett, Xiang Gao,
  Jianfeng Gao, Jingjing Liu, and Bill Dolan. 2019.
\newblock Dialogpt: Large-scale generative pre-training for conversational
  response generation.
\newblock \emph{arXiv preprint arXiv:1911.00536}.

\bibitem[{Zheng et~al.(2019)Zheng, Zhang, Mao, and
  Huang}]{DBLP:journals/corr/abs-1911-04700}
Yinhe Zheng, Rongsheng Zhang, Xiaoxi Mao, and Minlie Huang. 2019.
\newblock \href {http://arxiv.org/abs/1911.04700} {A pre-training based
  personalized dialogue generation model with persona-sparse data}.
\newblock \emph{CoRR}, abs/1911.04700.

\bibitem[{Zou et~al.(2021)Zou, Liu, Hu, and Zhang}]{zou-etal-2021-thinking}
Yicheng Zou, Zhihua Liu, Xingwu Hu, and Qi~Zhang. 2021.
\newblock \href {https://doi.org/10.18653/v1/2021.emnlp-main.169} {Thinking
  clearly, talking fast: Concept-guided non-autoregressive generation for
  open-domain dialogue systems}.
\newblock In \emph{Proceedings of the 2021 Conference on Empirical Methods in
  Natural Language Processing}, pages 2215--2226, Online and Punta Cana,
  Dominican Republic. Association for Computational Linguistics.

\end{thebibliography}

\clearpage

\appendix

\section{Additional Indicator Descriptions}
\label{sec:appendix_A}
We use PPL in our automatic evaluation metric for experimental verification. Since our method does not change models, the PPL of models does not change. 
Thus we choose GPT2 and BERT as language models and input the response into them to calculate PPL. Since the vocabulary of BERT is small, rare words generated by dialogue models may not be in the vocabulary of BERT, neither the baselines nor our method. This will cause the PPL to be huge. So when we use BERT to calculate PPL, the response with PPL greater than 10,000 are removed, both the response generated baselines and our method. Due to the vocabulary of GPT2 being large, the above operations are not required. 

\section{The Experimental Results of Speed}
\label{sec:appendix}
We test the generation speed of our method and baselines, and the results are shown in Table~\ref{table_A1}. The speed in Table~\ref{table_A1} is the time required to generate one response. All the generation speed is tested via PyTorch on 4 32GB NVIDIA V100 GPUs. As shown in Table~\ref{table_A1}, the generation speed of our method is faster than MMI, but slower than LLS. Although the LLS method is fast, the generation effect of LLS is significantly worse than our method as shown in Table~\ref{table1}. Furthermore, the performance of LLS is also significantly lower than that of MMI. Our method mainly brings additional time cost in the over-sampling stage, and the time cost in the post-evaluation stage is small. However, MMI takes a lot of time in both the generation and evaluation stages. It also proves that it is reasonable for us to use the TF-IDF algorithm instead of some learnable coherence evaluation method in the post-evaluation stage.

In addition, one of the reasons why BoB is generated significantly slower than Multi-GPT2 is that BoB has two decoders. The first decoder generates a preliminary response, and the second decoder modifies the preliminary response and generates the final response. Thus BoB generates two responses each time. Furthermore, the compression and distillation models effectively speed up our method.

\begin{table}[htp]
\centering
\scalebox{0.72}{
\begin{tabular}{lccc}
\hline
\textbf{} & \textbf{Generation} & \textbf{Evaluation} & \textbf{Sum} \\
\hline
BoB + MMI & 69.4s & 9.9s & 79.3s \\
BoB + LLS & 0.5s & - & 0.5s \\
BoB + SimOAP & 69.4s & 1.5s & 70.9s \\
BoB$_\mathrm{medium}$ + SimOAP & 23.7s & 1.4s & 25.1s \\

\hline
\hline
Multi-GPT2 + MMI & 10.1s & 10.1s & 20.2s \\
Multi-GPT2 + LLS & 0.1s & - & 0.1s\\
Multi-GPT2 + SimOAP & 10.1s & 1.3s & 11.4s \\
Multi-GPT2$_\mathrm{distil}$ + SimOAP & 5.8s & 1.3s & 7.1s \\
\hline
\end{tabular}
}
\caption{\label{table_A1}
The generation time of our method and baselines, the generation time includes two parts: response generation time (\textbf{Generation}) and response evaluation time (\textbf{Evaluation}).
}
\end{table}

\begin{table*}[htp]
\centering
\scalebox{0.97}{
\begin{tabular}{lccccc|c}
\hline
\textbf{} & \textbf{PPL}$_\mathrm{BERT}$ $\downarrow$ & \textbf{PPL}$_\mathrm{GPT2}$ $\downarrow$ & \textbf{Dis-1} $\uparrow$ & \textbf{Dis-2} $\uparrow$ & \textbf{C} $\uparrow$ & \textbf{Time} \\
\hline
BoB $w$ QuantiDCE & 15.58 &  80.03 & \textbf{16.25} & \textbf{45.78} & 0.456 & 7.4s  \\
BoB $w$ TF-IDF & \textbf{10.50} & \textbf{70.76} & 14.89 & 44.22 & \textbf{0.580} & \textbf{1.3s} \\
\hline
\hline
Multi-GPT2 $w$ QuantiDCE & 141.13 & 517.75 & \textbf{14.89} & \textbf{57.90} & 0.744 & 7.1s \\
Multi-GPT2 $w$ TF-IDF & \textbf{79.83} & \textbf{244.62} & 13.76 & 54.53 & \textbf{0.822} & \textbf{1.1s}\\
\hline
\end{tabular}
}
\caption{\label{table_A2}
Automatic evaluation results of SimOAP using QuantiDCE or TF-IDF.}
\end{table*}

\section{Further Analysis of Post-evaluation}
\label{sec:appendix_C}
To further analyze the method selection in the over-sampling stage of our method, we choose a learnable coherence evaluation method Quantifiable Dialogue Coherence Evaluation (QuantiDCE) \cite{ye-etal-2021-towards-quantifiable} to compare with TF-IDF. QuantiDCE trains a quantifiable coherence metric to reflect the actual human rating standards. QuantiDCE consists of multi-Level ranking pre-training and knowledge distillation fine-tuning. QuantiDCE uses BERT as a feature extraction module to encode the input context-response pair and then inputs the encoded features into a multi-layer perceptron (MLP) to obtain the final coherence evaluation score.

We use 500 pieces of data from the Personachat dataset for experimental validation. We first use the backbone models to generate 2,000 candidate responses each for the 500 pieces of data. Then QuantiDCE or TF-IDF is used to evaluate the coherence of the responses and select the 100 most coherent responses for each piece of data. Finally, the same natural language inference model is used to select the final response.

As shown in Table~\ref{table_A2}, coherence evaluation in the over-sampling stage using QuantiDCE outperforms TF-IDF on diversity. However, it is worse than TF-IDF in all other indicators. At the same time, the speed of QuantiDCE is much slower than TF-IDF. It is worth noting that for Multi-GPT2, the evaluation time cost of QuantiDCE is close to or even exceeds the time cost required by Multi-GPT2 in the oversampling phase in Table~\ref{table_A1}. For BoB, the evaluation time cost of QuantiDCE is more than 31\% of the over-sampling stage of BoB based on BERT-medium. Such evaluation time cost is unacceptable and avoidable. Combining the above two reasons, we chose fast and effective TF-IDF rather than other learnable methods in the coherence evaluation of the post-evaluation stage. 

After the coherence assessment in the post-evaluation, 100 highly coherent responses among 2000 candidates responses are selected. In the subsequent consistency evaluation, we use the natural language inference model to evaluate the consistency of 100 candidate responses. Although the evaluation speed of the natural language inference model is also slow, there are only 100 candidate responses to be evaluated for each dialogue at this time, and the time cost of this process is small, as shown in Table~\ref{table_A1}. At the same time, the natural language inference dataset DialogueNLI we use is specially built for persona-based dialogue. Many previous works on persona-based dialogue generation have also verified that it works well \cite{kim-etal-2020-will,DBLP:journals/corr/abs-1911-05889,song-etal-2020-generate,cao-etal-2022-model}. So we chose the natural language inference model fine-tuned on DialogueNLI in the consistency evaluation of the post-evaluation stage. 

\end{document}